\documentclass[acmtog]{acmart} 
\acmSubmissionID{806}

\usepackage{booktabs} 

\usepackage{microtype}
\usepackage{graphicx}
\usepackage{subfigure}
\usepackage{caption}
\usepackage{bbm}
\usepackage{url}
\usepackage{multirow,makecell, rotating}
\usepackage{tikz}
\usepackage{graphbox}
\usepackage{comment}
\usepackage{epsfig}
\usepackage{bm}
\usepackage{soul}
\usepackage{color}
\usepackage{hyperref}

\usepackage{amsmath}
\usepackage{mathtools}
\usepackage{amsthm}

\usepackage[ruled]{algorithm2e} 

\SetAlFnt{\small}
\SetAlCapFnt{\small}
\SetAlCapNameFnt{\small}
\SetAlCapHSkip{0pt}

\usepackage[capitalize]{cleveref}
\crefname{section}{Sec.}{Secs.}
\Crefname{section}{Section}{Sections}
\Crefname{table}{Table}{Tables}
\crefname{table}{Tab.}{Tabs.}

\setcopyright{rightsretained}
\acmJournal{TOG}
\acmYear{2023} \acmVolume{42} \acmNumber{4} \acmArticle{1}
\acmMonth{8} \acmPrice{15.00}\acmDOI{10.1145/3592451}

\begin{document}
\title{UniTune: Text-Driven Image Editing by Fine Tuning a Diffusion Model on a Single Image}

\author{Dani Valevski}
\orcid{0000-0002-4211-2866}
\affiliation{%
  \institution{Google Research}
  \country{Israel}}
\email{daniv@google.com}
\authornote{Equal contribution.}
\author{Matan Kalman}
\orcid{0009-0001-8002-1148}
\affiliation{%
  \institution{Google Research}
  \country{USA}}
\email{matank@google.com}
\authornotemark[1]
\author{Eyal Molad}
\orcid{0009-0005-4353-9713}
\affiliation{%
  \institution{Google Research}
  \country{Israel}}
\email{moladeyal@google.com}
\author{Eyal Segalis}
\orcid{0009-0004-6529-5361}
\affiliation{%
  \institution{Google Research}
  \country{Israel}}
\email{eyalis@google.com}
\author{Yossi Matias}
\orcid{0000-0003-3960-6002}
\affiliation{%
  \institution{Google Research}
  \country{Israel}}
\email{yossi@google.com}
\author{Yaniv Leviathan}
\orcid{0009-0000-4080-4845}
\affiliation{%
  \institution{Google Research}
  \country{USA}}
\email{leviathan@google.com}
\authornotemark[1]

\begin{abstract}
Text-driven image generation methods have shown impressive results recently, allowing casual users to generate high quality images by providing textual descriptions. However, similar capabilities for editing existing images are still out of reach. Text-driven image editing methods usually need edit masks, struggle with edits that require significant visual changes and cannot easily keep specific details of the edited portion. In this paper we make the observation that image-generation models can be converted to image-editing models simply by fine-tuning them on a single image.
We also show that initializing the stochastic sampler with a noised version of the base image before the sampling and interpolating relevant details from the base image after sampling further increase the quality of the edit operation.
Combining these observations, we propose UniTune, a novel image editing method. UniTune gets as input an arbitrary image and a textual edit description, and carries out the edit while maintaining high fidelity to the input image. UniTune does not require additional inputs, like masks or sketches, and can perform multiple edits on the same image without retraining. We test our method using the Imagen model in a range of different use cases. We demonstrate
that it is broadly applicable and can perform a surprisingly wide range of expressive editing operations, including those requiring significant visual changes that were previously impossible.

\end{abstract}

\begin{teaserfigure}
\centering
\includegraphics[width=0.80\textwidth]{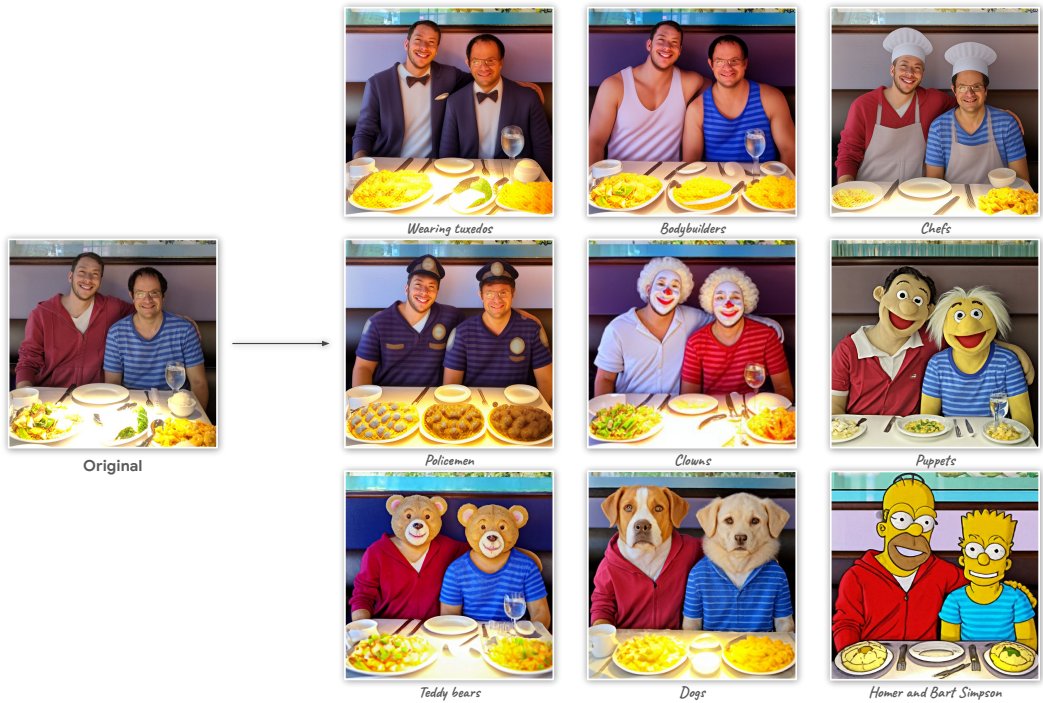}
    \vspace{-0.2cm}
    \caption{
   UniTune edits preserving visual and semantic fidelity to the user supplied image.
    }
    \label{fig:edit_variety}
\end{teaserfigure}

\maketitle

\begin{figure*}[ht]
\centering
\includegraphics[width=.8\linewidth]{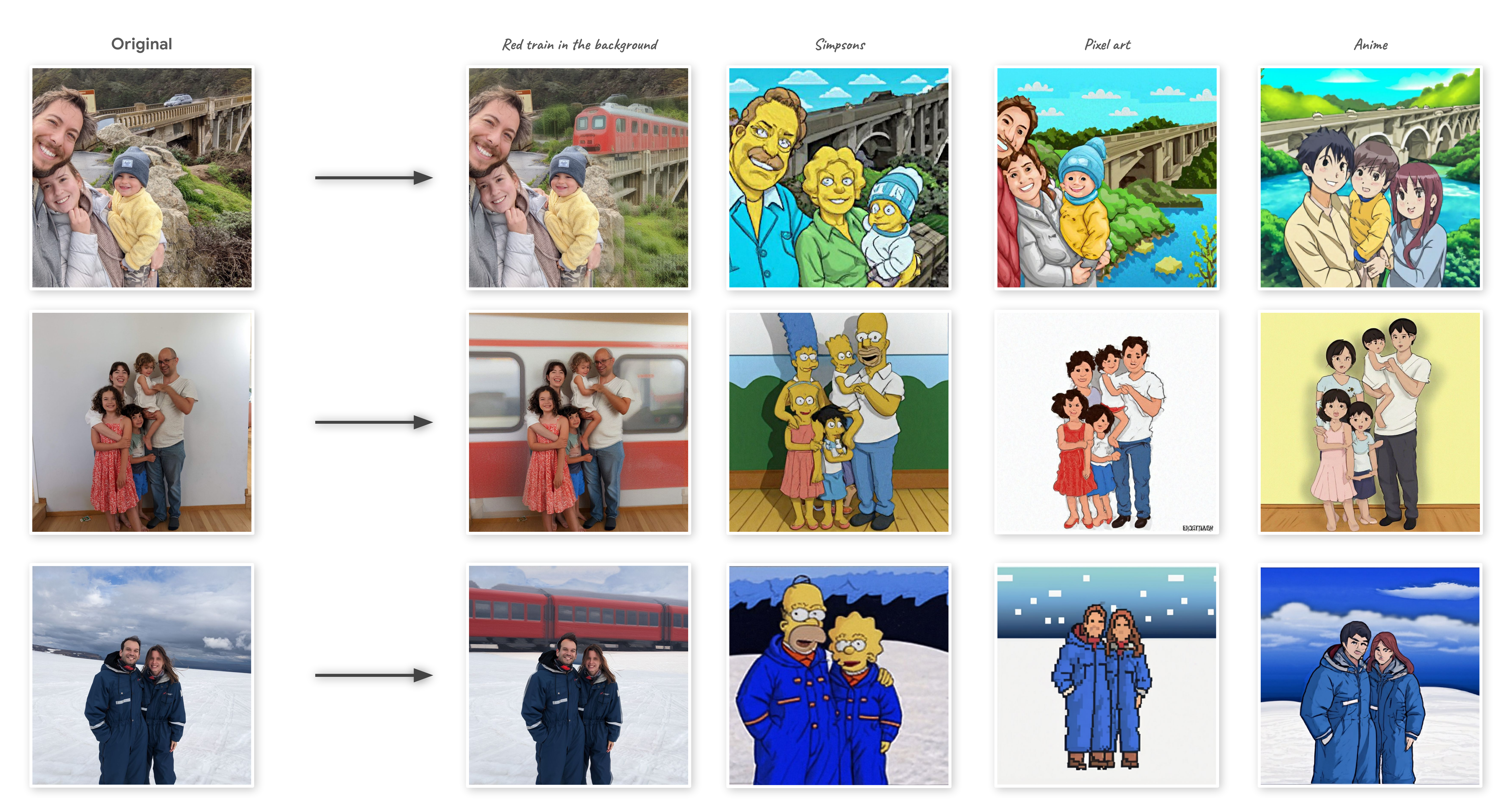}
\caption[fig2]{UniTune maintains high visual fidelity in unaffected areas of the image when making localized edits (e.g. the faces when adding a train to the background), and high semantic fidelity when making global edits (e.g. when changing the style of the entire image).}
\label{fig:families}
\end{figure*}

\section{Introduction}
High fidelity image manipulation via text commands is a long standing problem in computer graphics research. Using free-form commands to describe a desired edit, like “men wearing tuxedos”, “pixel art”, or “a blue house” (\cref{fig:edit_variety,fig:families,fig:house}) is significantly easier than carrying out the changes manually in an image editing software. Intuitive language based interfaces have the potential to make experts more efficient and to unlock graphic design capabilities for casual users. Despite amazing advancements in image generation methods \cite{DallE2, Imagen}, general domain high-fidelity image editing is still an unsolved problem. In this work, we present UniTune, which takes a meaningful step towards that goal.

Revolutionary text-to-image models like Dall-E \cite{DallE2}, Imagen \cite{Imagen, sr3} and Stable Diffusion \cite{StableDiffusion} excel at creating images from scratch, or at filling manually-removed parts of existing images in a context-aware fashion. However, for editing operations, these models usually require the user to specify masks and often struggle with edits that depend on the masked portion of the image.

UniTune is a novel method to edit images by simply supplying a textual description of the desired result preserving high fidelity to the entirety of the input image, including the edited portions. Fidelity is preserved both to visual details (e.g. shapes, colors, and textures) and semantic details (e.g. objects, poses, and actions) as required by the edit operation.

UniTune is able to edit arbitrary images in complex cross domain scenes. We tested it for localized edits as well as broad global edits (see \cref{sec:Results}). To our knowledge, UniTune is unique in its ability to pick the correct details to preserve, allowing it for example to change clothes, positions and actions without changing characters, or create a caricature of an image that maintains semantic details like posture and hair style.

UniTune performs expressive image editing by harnessing the power of large scale text-to-image diffusion models. We show a simple yet powerful technique for transferring their visual and semantic capabilities to the domain of image editing. 

\begin{figure}[h]
\centering

\includegraphics[width=1\linewidth]{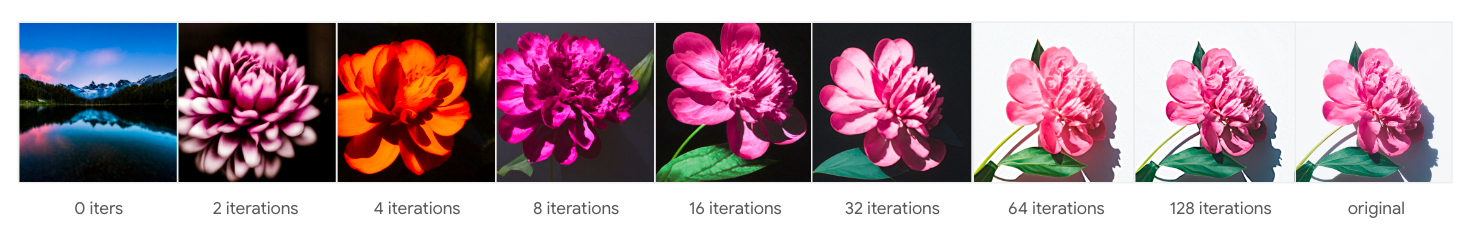}
\caption{The images generated by the model conditioned on the tokens it was trained on after different number of fine tuning iterations. It takes around 64 iterations for the model to be able to faithfully reproduce the original image. Input source: Unsplash.}
\label{fig:fine_tune_iterations}
\end{figure}

\begin{figure}[h]
\centering
\includegraphics[width=1\linewidth]{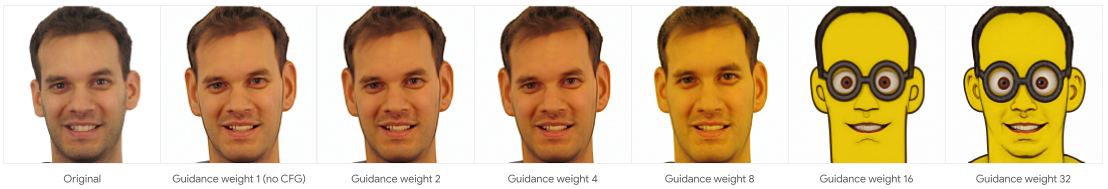}
\caption[test]{Images for the prompt "minion" after fine-tuning on the image on the left, with various degrees of Classifier Free Guidance weights. With standard conditioned sampling (second from the left), the model is biased towards the image it was fine-tuned on. Using Classifier Free Guidance (right) we observe that the knowledge of what a "minion" is, is preserved within the model's weights.}
\label{fig:cfg}
\end{figure}

Our main observation is that, with the right parameters, fine-tuning large diffusion models on a single $(image,prompt)$ pair does not lead to \emph{complete} catastrophic forgetting \cite{CatastrophicForgetting}. As expected, a fine-tuned model will strongly prefer to associate the provided image and prompt together (see \cref{fig:fine_tune_iterations}), and will strongly prefer to draw samples that are almost identical to the provided image given other prompts. However, the visual and semantic knowledge that the model acquired in its original training is still usable across a very wide variety of edit operations when using Classifier Free Guidance (see \cref{fig:cfg}). The fidelity-expressiveness balance can be tuned by controlling the number of fine tuning iterations, the weights of Classifier Free Guidance \cite{ClassifierFreeGuidance} or by replacing a number of initial sampling steps with a noisy version of the base image (as in \cite{SDEdit}).

Fine-tuning of diffusion models is a powerful technique, relevant to many use cases like image-to-image translation \cite{wang2022pretraining} and topic-driven image generation \cite{Dreambooth, TextualInversion}. These approaches attempt to mitigate over-fitting at training time by data augmentation \cite{Dreambooth}, using large data sets \cite{wang2022pretraining, Glide} or limiting fine-tuning to the embedding of specific tokens \cite{TextualInversion}. This allows these techniques to learn e.g. the essence of a subject, without learning transient image-specific attributes, like pose, camera angle, background, etc. For our use case of image editing, some over-fitting is beneficial as we in fact aim to maintain high fidelity to the source image.

\section{Related work}
\label{sec:RelatedWork}
Image editing is a fundamental problem in computer graphics research and finding intuitive interfaces for image manipulation has been an active research field for years.

\textbf{GAN based editors.} Some solutions turned GANs \cite{GAN, StyleGAN} into image editors by inverting the base image into the GAN latent space \cite{GanInversionSurvey, GenerativeVisualManipulation, NeuralPhotoEditing, PivotalTuning} and use CLIP \cite{CLIP} to guide the generator \cite{StyleClip, TediGAN, PaintByWord, Clip2StyleGAN, ConditionalImageGenerationandManipulation, stylegannada}. These methods achieve impressive results given a well organized latent space like StyleGan \cite{StyleGAN} but are limited to the the domains where the GAN model is trained (e.g. faces).

Diffusion models \cite{NonequilibriumThermodynamics, DDPM, EstimatingGradients} began to outperform GANs \cite{DiffusionModelsBeatGan} and are used by newer image editing methods (especially for multi-domain use cases).

\textbf{Sampling based fidelity techniques.} For image fidelity, diffusion models self-correcting sampling is extremely useful. DiffusionClip \cite{DiffusionClip} uses a noising deterministic DDIM process to find the exact noise that will result in the target image. ILVR \cite{ILVR} and Liu et al. \cite{MoreControl} use classifier Guidance \cite{DiffusionModelsBeatGan} to guide a DDPM sampler in a direction that is close to the base image, visually or semantically. These methods were demonstrated in narrow domains (e.g. single object) or with limited edit operations.

\textbf{SDEdit} \cite{SDEdit} performs edits by starting the sampling process from a noisy version of the base image. It is a powerful method and is used for image editing in Stable Diffusion img2img mode. However, since the model has never seen the base image, the denoising process cannot recover finer details from the original image. Moreover, the method is heavily biased towards the pixel values in the original image, making it unsuitable for significant pixel-level changes.

\textbf{Edit Masks.}
To address the problem of picking the correct details to preserve, methods like Blended Latent Diffusion \cite{BlendedDiffusion, BlendedLatentDiffusion} and Glide \cite{Glide} use edit masks which limit the edit to a specific area of the image. However, the detail selection problem still exist within the edited portion and it's sometimes hard to predict the size of the desired edit area.

\textbf{Prompt-to-Prompt} \cite{Prompt2Prompt} achieves high quality editing capabilities by operating on the outputs of a text-to-image model. This allows using visual-semantic information encoded in its intermediate attention matrices. However, relying on the attention weights means the method works only on images that were generated by the diffusion model (they experimented with inverting arbitrary images using DDIM noising with mixed results), requires a prompt that describes the base image, and cannot change the location of objects.

\textbf{Text2LIVE} \cite{text2live} trains a U-Net \cite{UNet} on the fly to perform the edit operation on the base image. The training data is the image itself (in different crops and augmentations), and the loss function is CLIP based. The model excels on local, pixel structure preserving changes like adding effects (fire, smoke) or changing texture and color, but cannot perform complex edits.

\textbf{Subject-driven image generation.} Two recent breakthroughs, DreamBooth \cite{Dreambooth} and Textual Inversion \cite{TextualInversion} solve a related but distinct problem of subject-driven image generation. In that setting the goal is to use a small number of images to teach a text-to-image model to generate novel renditions of a given subject in new contexts. The model should derive the details that are important to the subject, but avoid keeping details like background, position, and other objects in the images (as opposed to UniTune). Both methods use fine-tuning (either of the image generation model or of an entry in the embedding table of the textual encoder).

\textbf{Concurrent methods.} Several related methods were developed concurrently and independently with our work. DreamArtist \cite{dreamartist} can edit images, but is mainly aimed at subject driven generation. InstructPix2Pix \cite{brooks2023instructpix2pix} targets image editing directly, and does so by training a targeted model for image editing based on an augmented dataset. Most related to UniTune and sharing the core idea, Imagic \cite{imagic} also edits images by fine tuning a diffusion image generation model on a single image. One difference is that Imagic fine-tunes the base model for every edit-prompt, whereas UniTune only fine-tunes once per base image, making subsequent edits faster.

\section{Method}
\label{sec:method}
Our goal is to convert a text-to-image diffusion model, $f_\theta$ into a text-conditioned image editor, $f_{\theta^*}$ for a specific user provided base image, $x^{(b)}$. The new model should be able to accept edit prompts $c$, that describe the image after the edit (see appendix for examples of edit prompts), and output an edited image, $x_0$ that satisfies the condition $c$ and maintains fidelity to $x^{(b)}$.

Overall our method is composed of two stages: (1) fine-tune the model on the base image $x^{(b)}$ alone (2) use a modified sampling process that balances fidelity to the base image $x^{(b)}$ and alignment to the edit prompt $c$. 

\subsection{Fine-tuning}
\label{sec:method-fine-tuning}
We fine-tune the model on $x^{(b)}$ for a fixed number of steps, encouraging it to produce images that are close to the base image. Following \cite{Dreambooth}, we use a text-condition during the fine-tuning stage, $c^{(b)}$, that is composed of 3 rare tokens, creating a rare word which is not found in the original training data of $f_\theta$. We use the diffusion model denoising loss with fixed condition and image:

\begin{equation}
  \mathbb{E}_{\boldsymbol\epsilon,t}[w_t||\boldsymbol{f}_\theta(\alpha_t\boldsymbol{x^{(b)}}+\sigma_t\boldsymbol\epsilon,\boldsymbol{c^{(b)}})-\boldsymbol{\epsilon}||^2_2]
  \label{eq:important}
\end{equation}

where $t\sim\mathcal{U}([0,1])$, $\boldsymbol\epsilon\sim\mathcal{N}(0,I)$ and $w_t$, $\alpha_t$, $\sigma_t$ are functions of $t$ determined by the noising schedule of the diffusion model (see \cite{DDPM} for more details).

\begin{figure}
\includegraphics[width=1\linewidth]{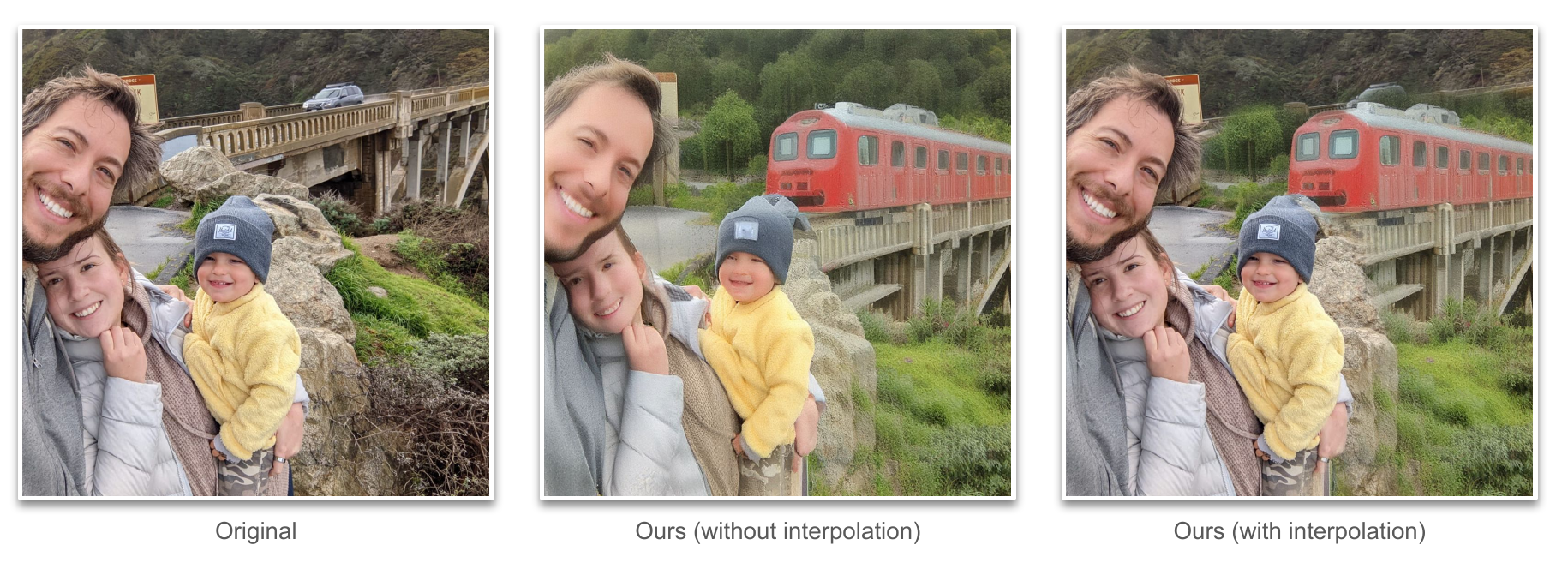}
\caption{An example showing the benefit of interpolation to quality.}
\centering
\label{fig:interpolation}
\end{figure}

\subsection{Sampling}
To perform the edit operation, we sample the fine-tuned model by concatenating $c$ and $c^{(b)}$ (i.e., the string "[$rare\_tokens]\ edit\_prompt$"). With naive sampling, the fine-tuned model's bias towards $x^{(b)}$ outweighs the provided prompt $c$ and the model produces an image very similar to $x^{(b)}$. Classifier free guidance \cite{ClassifierFreeGuidance} is used to guide the model towards the concatenated prompt (\cref{fig:cfg}) producing an image that maintains fidelity to $x^{(b)}$ while satisfying $c$. Since we use a high value for the Classifier Free Guidance weight we apply Oscillating Guidance \cite{ImagenVideo} and Dynamic Thresholding \cite{Imagen}. To increase visual fidelity, we begin sampling at a lower step $t$ (instead of starting with $t=1$) and initialize the sampling with an appropriately noised version of $x^{(b)}$ (instead of random Gaussian noise) using the diffusion forward process, following \cite{SDEdit}:

\begin{equation}
  \boldsymbol{z_t} = \alpha_t\boldsymbol{x^{(b)}}+\sigma_t\boldsymbol\epsilon
\end{equation}
$\boldsymbol{z_t}$ is the initialization value, $\boldsymbol\epsilon\sim\mathcal{N}(0,I)$ and $\alpha_t$, $\sigma_t$ are functions of $t$ determined by the noising schedule of the diffusion model. Finally, to further preserve fine details from the source image $x^{(b)}$, we linearly interpolate the pixels of the generated image with the pixels of $x^{(b)}$. The interpolation weight is determined by the similarity of the pixel neighborhoods\footnote{Neighborhood similarity is the result of applying a transformed Gaussian filter to the sum across channels of the squared distances between the original image $x^{(b)}$ and generated image.} (see \cref{fig:interpolation}).

\subsection{Details}
In the experiments we performed, we used Imagen \cite{Imagen} as the text-to-image model, with a frozen T5-XXL encoder for text embedding \cite{t5} (see appendix for details of an implementation on top of Stable Diffusion \cite{StableDiffusion}). Imagen is composed of a text-to-image model that generates 64x64 pixels output, and two super resolutions models that convert the 64x64 image to a 256x256 image and then to a 1024x1024 image. We fine-tune the first two models as described in \cref{sec:method-fine-tuning} and use the default 1024 model. We train the 64x64 model with Adafactor and the 256x256 model with Adamw  with a learning rate of 0.0001 (the same setting used when training Imagen). We use a batch size of 4 and emit weights at 16, 32, 64, 128 training steps (we use less training steps when expressiveness is needed, and more when fidelity is needed). In this setting, our fine-tuned model can reproduce $x^{(b)}$ after 64 iterations (\cref{fig:fine_tune_iterations}). 

We observe that when fine-tuning with a very low number of steps (16-128) and using Classifier Free Guidance, the model takes into account the edit prompt and then returns an image $x_0$ that is similar to $x^{(b)}$ and satisfies $c$ as desired. Surprisingly, despite the fact that we used a pixel-level MSE loss at fine-tuning, the similarity of $x_0$ to $x^{(b)}$ is often semantic and their MSE can be very different (see \cref{sec:Results}), hinting that fine-tuning changed the biases in an internal semantic representation of the model.

In our sampling experiments we discover that when using a fine-tuned model, starting the sampling with very noisy versions of $x^{(b)}$ is enough to maintain high visual fidelity (because the model is already biased towards $x^{(b)}$). Therefore, to increase visual fidelity we start the sampling at step $0.8 \le t \le 0.98$ and noise the base image appropriately. We also experimented with various values for Classifier Free Guidance, and picked a value of 32 which performed best. Additional details on the effect of different parameters on the results are provided in the appendix.

\begin{figure*}
\includegraphics[width=1\linewidth]{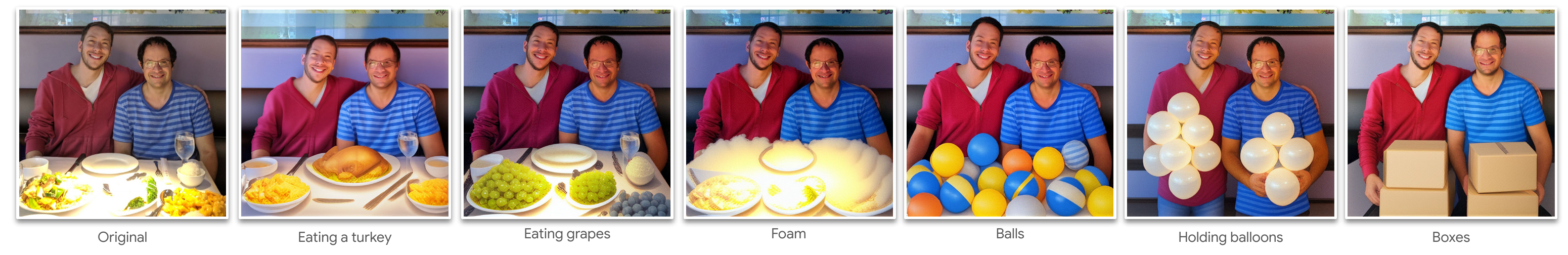}
\caption[test]{Examples of UniTune performing local edits (adding objects) without masks.}
\centering
\label{fig:local_edits}
\end{figure*}

\section{Results}
\label{sec:Results}
\begin{figure*}[htbp]
\centering
\includegraphics[width=0.75\linewidth]{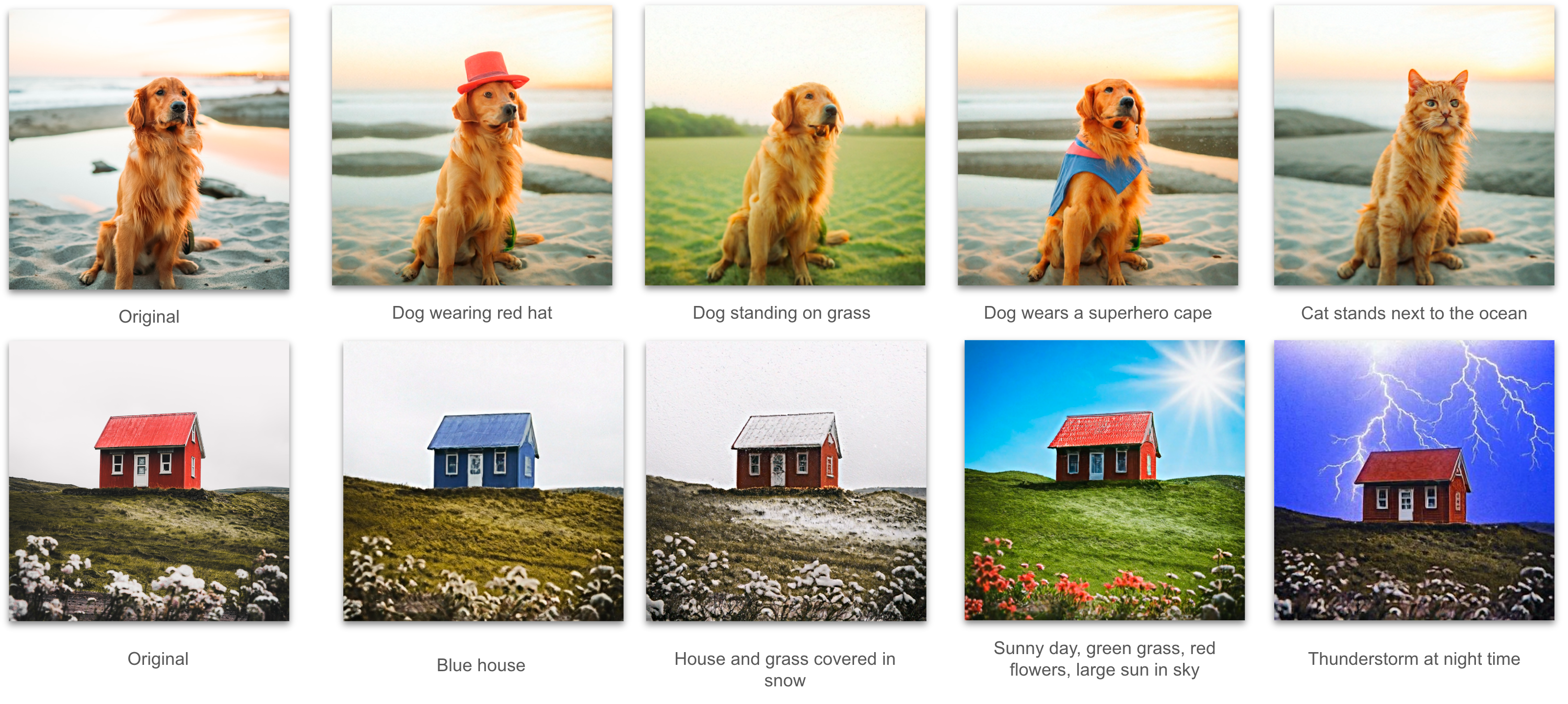}
\caption[test]{UniTune works in multiple domains and can carry out a combination of local and global manipulations. Input source: Unsplash.}
\label{fig:house}
\end{figure*}

\begin{figure*}[htbp]
\centering
\includegraphics[width=0.75\linewidth]{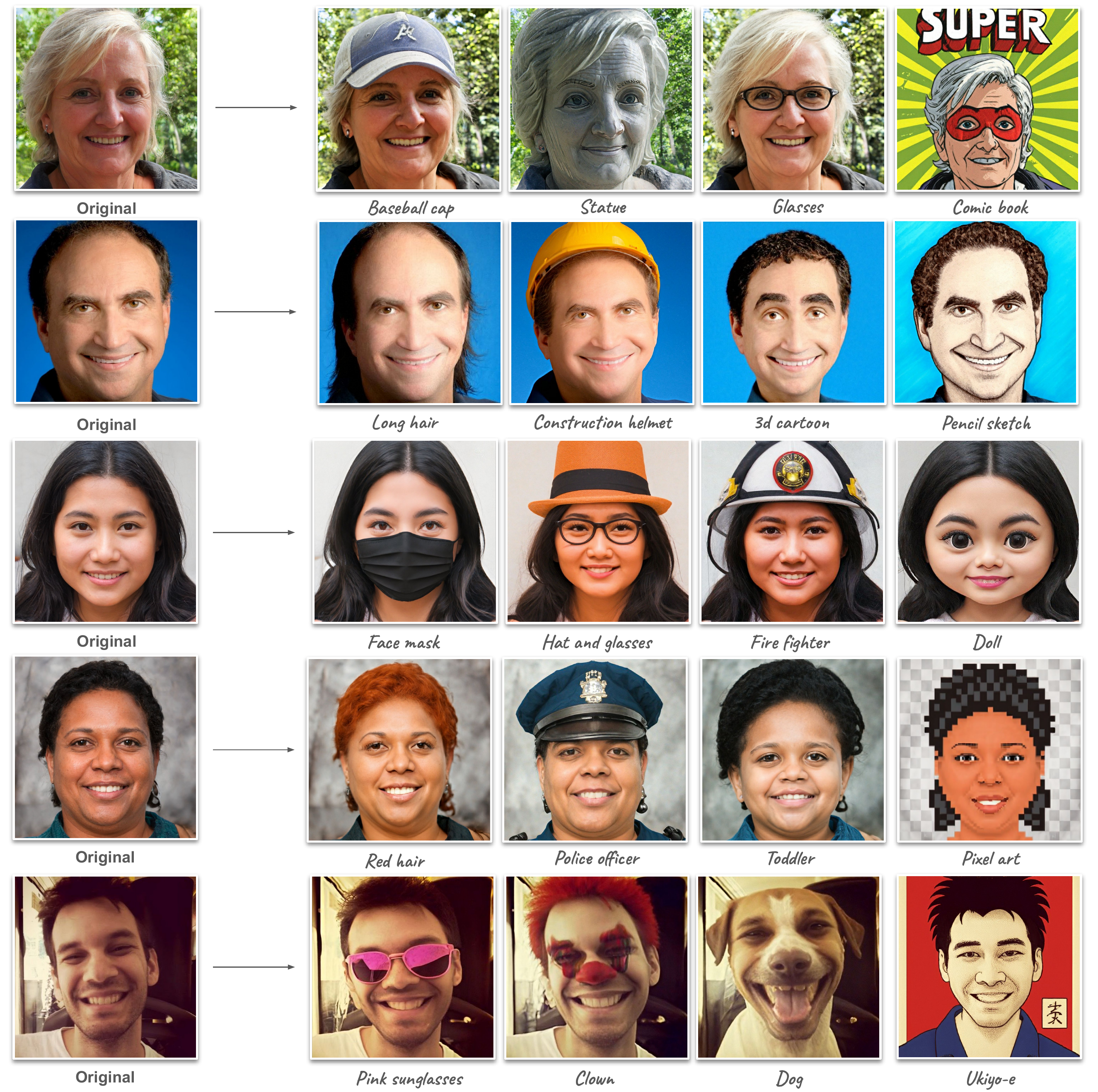}
\caption[test]{UniTune shows comparable editing capabilities to models trained on a single domain like portrait photos.}
\label{fig:faces}
\end{figure*}

\begin{figure*}[htbp]
\centering
\includegraphics[width=0.75\linewidth]{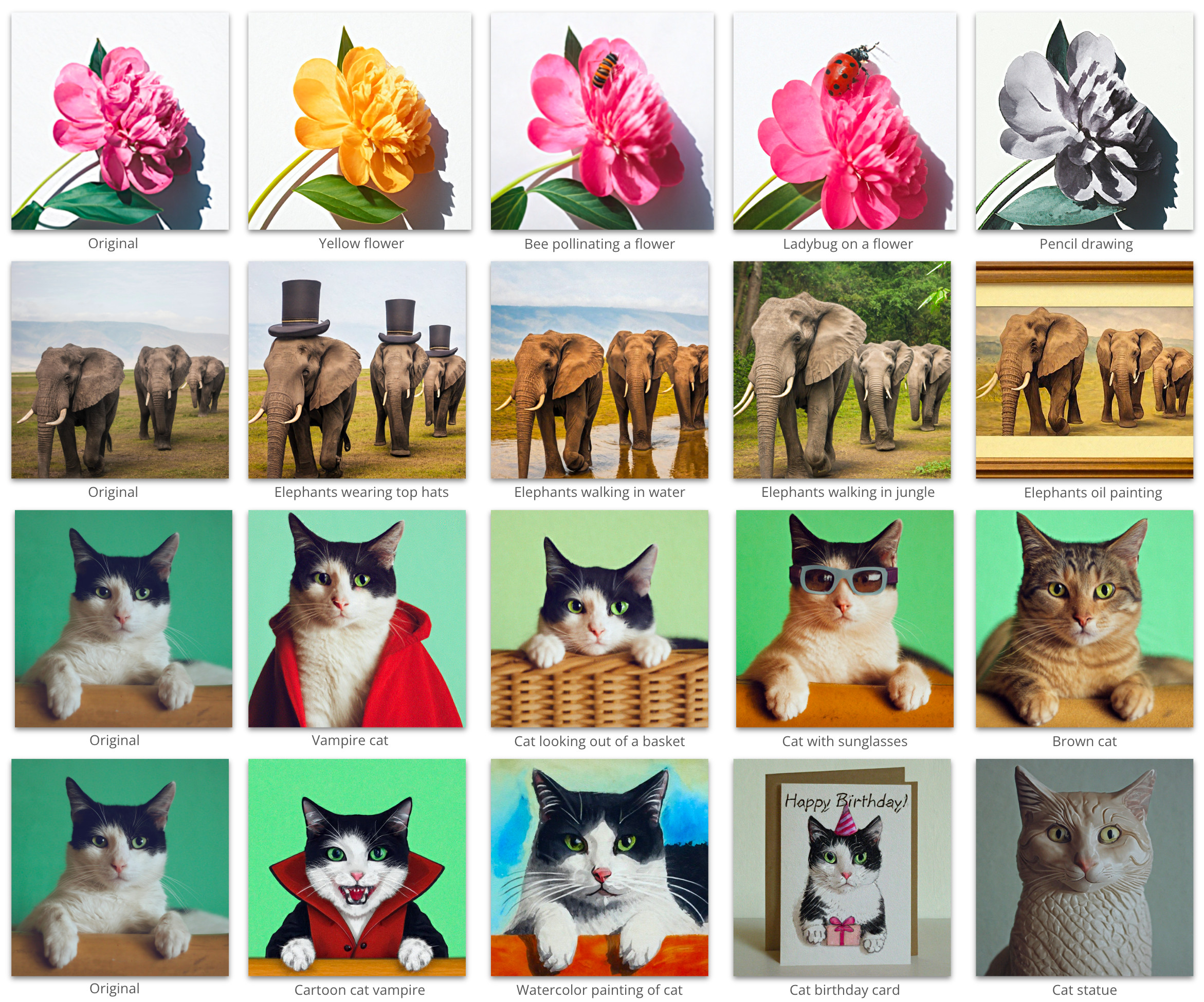}
\caption[test]{More UniTune examples. The bottom two rows demonstrate visual and semantic fidelity. See the appendix for the full edit prompts and parameter settings used to generate the images in this figure. Input source: Unsplash.}
\label{fig:cats}
\end{figure*}

To demonstrate UniTune's breadth of edit capabilities we tested the system in multiple scenarios, including adding or changing items in the scene, adding an accessory, changing hair-style or clothing, changing the scene's background, and changing the overall look and style of the photo (see \cref{fig:edit_variety,fig:families,fig:house,fig:faces,fig:cats,fig:local_edits}).

\textbf{Varying fidelity levels.} While all image-editing operations require some level of similarity to the base image, the desired level varies greatly between use cases. UniTune can make edits in a wide range of fidelity levels. \cref{fig:edit_variety,fig:local_edits} show how different prompts result in localized changes (e.g., adding a turkey to an empty plate), changes that alter accessories, clothing or surroundings but keep the character's visual appearance (e.g., changing plain clothes to tuxedos), changes in the characters themselves (e.g., changing people into dogs), and changes that transform the style of the image, keeping only semantic details intact (e.g., changing a realistic scene into a cartoon one).

\textbf{Semantic understanding of the base image.} \cref{fig:families} demonstrates a unique capability of UniTune: making changes that only maintain semantics and not necessarily pixel-level details. Images in the last two columns maintain semantic features like clothing, hairstyle, spatial orientation and weather while pixel level details are very different.

\textbf{Localized edits.} UniTune can also perform localized edits, with or without edit masks. \cref{fig:cats} shows how UniTune can place new objects (bee, ladybug, hats) without an edit mask. When edit masks are used, UniTune strength is in scenarios where familiarity with the the details under the mask is required, as demonstrated in the appendix. \cref{fig:edit_variety,fig:faces} show some examples of accessorizing, which is a type of local edit operation that is hard to achieve without familiarity with the details of the edited portion. 

\textbf{Narrow domains.} \cref{fig:faces,fig:house} demonstrates complex edits in a narrow domains. While domain specific methods (e.g. GANs trained on portrait photos) show strong performance in these cases, UniTune has comparable capabilities across numerous domains and can apply multiple changes simultaneously.

\textbf{Seed selection and hyperparameters.} Most of the figures in the paper were chosen out of 8 generations produced by different seeds and the initial prompt we tried. In the cases where all 8 generations were rejected, we experimented with more seeds, changing the hyperparameters, or with minimal prompt engineering. Usually, no more than a few attempts were needed.

\section{Comparisons}

We compare UniTune to several method qualitatively, showing comparable results on images from the relevant works. We also compare UniTune to SDEdit quantitatively, and show overall a significant preference for UniTune (72\% vs 28\%). We analyze these results, and observe that both methods perform well in cases where the edits don't require changing the pixels of the base image in a significant way, but UniTune performs much better in cases where many pixels needs to change significantly (e.g. when duplicating, moving or resizing objects).

\subsection{Qualitative Comparisons}
\begin{figure}[ht]
\includegraphics[width=1\linewidth]{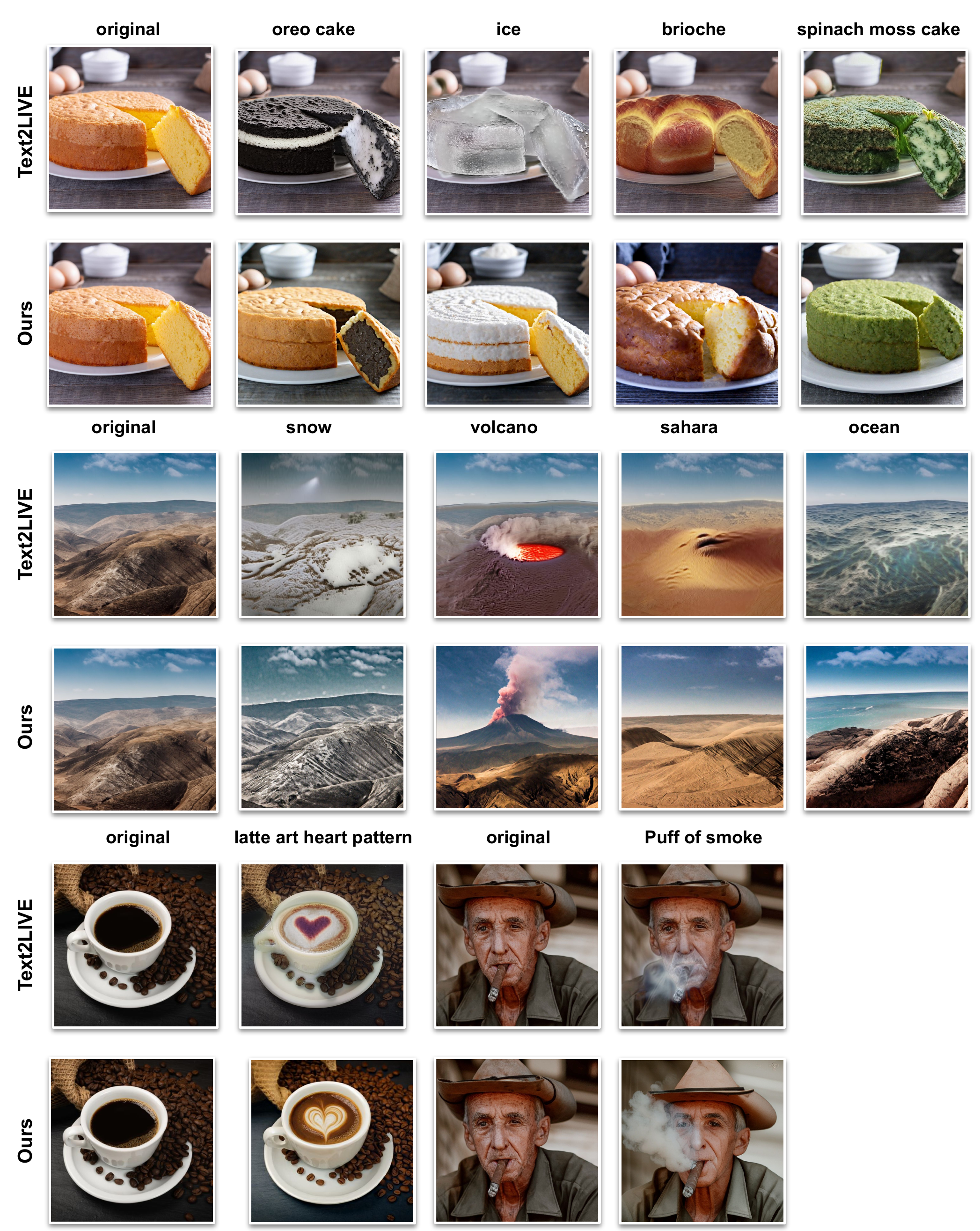}
\caption[test]{Comparison of UniTune and Text2LIVE. The original images and Text2LIVE edits are taken from the Text2LIVE paper \cite{text2live}. We observe comparable results for the types of edits Text2LIVE supports. We have not compared UniTune to Text2LIVE for complex edits, like adding new objects or deviating from the original layout, as Text2LIVE does not support these kinds of edits. To generate UniTune examples we manually selected the best result out of 64 variations generated with different UniTune configurations. The edit prompts for the last two examples were slightly changed from the original paper to mention 'coffee cup' and 'cigar'.}
\centering
\label{fig:text2live_comp_full}
\end{figure}

\begin{figure*}[ht]
\centering
\includegraphics[width=0.8\linewidth]{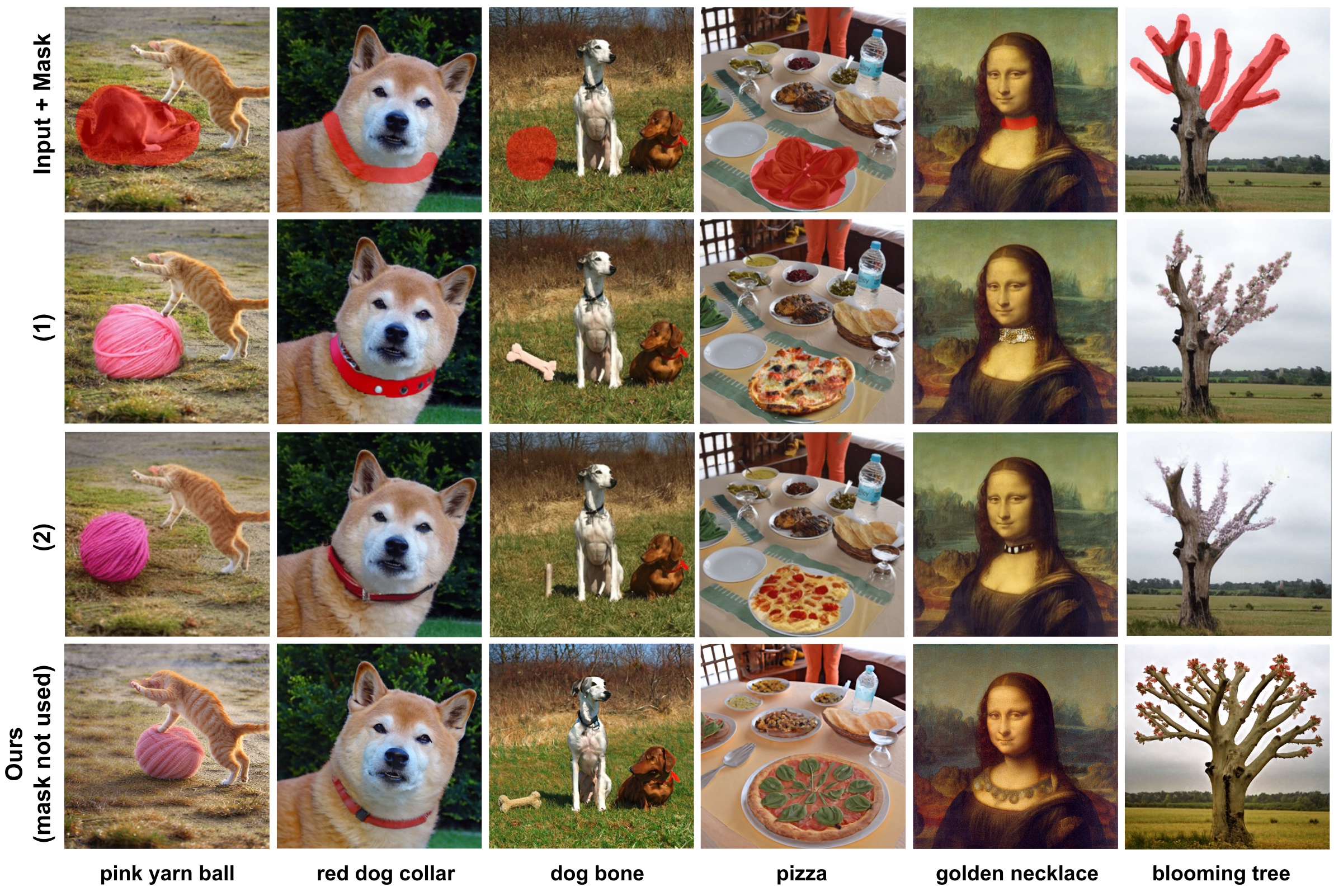}
\caption[test]{Comparison of image editing in UniTune to in-painting in Glide and Blended Latent Diffusion. (1) Glide (implicit) \cite{Glide} (2) Blended Latent Diffusion \cite{BlendedLatentDiffusion},  both using the best sample out of 64 selected by CLIP. The top three rows of the figure were taken from the papers above. Since no mask is supplied to UniTune, edit prompts were altered to pinpoint the correct part of the figure to change. To generate the UniTune examples, we tried a few edit prompt phrasings, and manually selected the best result out of 64 variations generated with different UniTune configurations.}
\label{fig:inpainting_comp}
\end{figure*}

We provide a qualitative comparisons of Unitune to various methods that work on arbitrary images. \cref{fig:text2live_comp_full} compares Unitune to Text2Live \cite{text2live}, a method that focuses on changing textures or adding effects like smoke, but not on modifying complex structures. \cref{fig:inpainting_comp} is a comparison of Unitune to various in-painting methods. We used Unitune without a mask in these examples, and changed the prompt to correctly pinpoint the exact part of the image we wanted to edit. As an example, "a cat jumping on a pink yarn ball" was used instead of "pink yarn ball", to guide UniTune towards selecting the correct cat to turn into a yarn ball.

\subsection{Quantitative comparisons}

\begin{table}
\centering
\caption{Preference by majority vote of UniTune vs. SDEdit, scoped to aligned an unaligned edits.}
\begin{tabular}{|c c c|} 
 \hline
 & UniTune & SDedit \\
 \hline
 Overall & 72\% & 28\% \\
 Unaligned & 83\% & 17\% \\ 
 Aligned & 60\% & 40\% \\ 
 \hline
\end{tabular}
\label{table:popular}
\end{table}

\begin{table}
\centering
\caption{Human evaluation of the top edited image per method, when scoped to aligned (A) and unaligned (U) edits.}

{
    \begin{tabular}{|c c c c| } 
     \hline
     Method & Quality  & Fidelity  & Text Align. \\
     \hline
     UniTune  & $ 4.14 \pm 0.75 $ & $ 3.87 \pm 0.75 $ & $ 4.21 \pm 0.78 $  \\ 
     SDedit  & $ 3.90 \pm 0.84 $ & $ 3.54 \pm 0.79 $ & $ 3.89 \pm 0.93 $ \\
     \hline
     UniTune (U)  & $ 4.09 \pm 0.70 $ & $ 3.92 \pm 0.76 $ & $ 4.29 \pm 0.75 $  \\ 
     SDedit (U)  & $ 3.68 \pm 0.81 $ & $ 3.38 \pm 0.79 $ & $ 3.68 \pm 0.99 $ \\
     \hline
     UniTune (A) & $ 4.19 \pm 0.79 $ & $ 3.81 \pm 0.74 $ & $ 4.12 \pm 0.81 $ \\
     SDedit (A) & $ 4.13 \pm 0.80$ & $ 3.71 \pm 0.76 $ & $ 4.11 \pm 0.80 $ \\ 
     \hline
    \end{tabular}
}
\label{table:questions}
\end{table}

\begin{figure}
\includegraphics[width=1\linewidth]{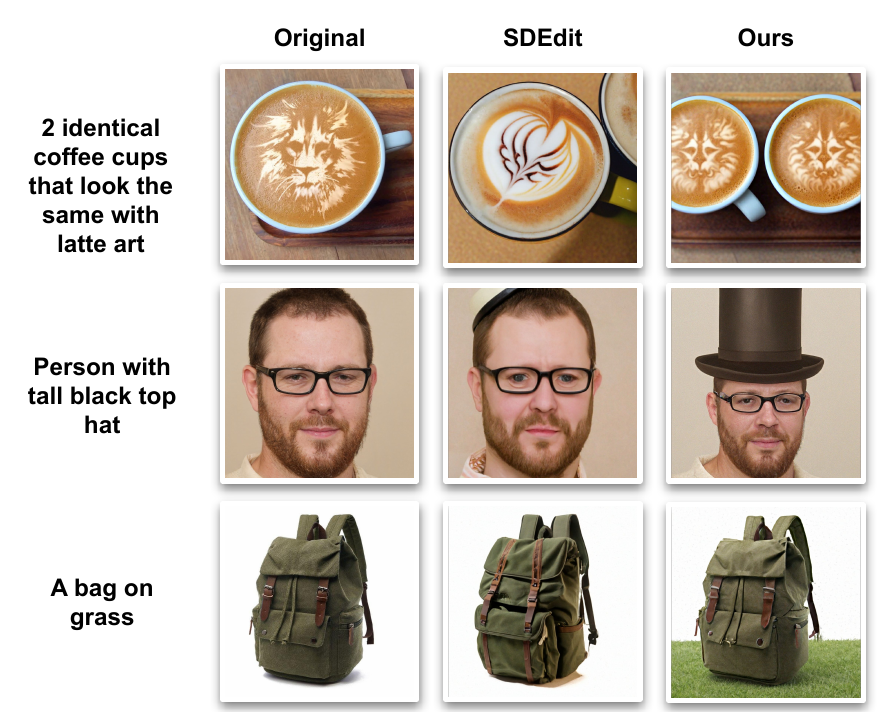}
\caption[test]{Examples of SDEdit and UniTune performance on edits that require significant pixel-level visual changes ("unaligned edits"). Input source for rows 1, 3: Unsplash.}
\centering
\label{fig:unaligned}
\end{figure}

We performed a quantitative comparison between UniTune and SDedit \cite{SDEdit} with a panel of 29 raters (\cref{table:popular,table:questions}). Though we expect both methods to be used in a complementary fashion, SDEdit serves as a good baseline since it works on arbitrary images and can perform edits with no masks. The comparison was done on a dataset of 93 (base image, textual edit) pairs with images of animals, people, objects, food and scenery. 51\% of the edits in the dataset required significant pixel-level visual changes (like moving objects, zooming out or making significant style changes, see \cref{fig:unaligned}). We refer to this class of edits as "unaligned".

\cref{table:popular} summarizes an evaluation of the overall preferred method and \cref{table:questions} summarizes evaluation of the chosen edited image per method on 3 attributes. Additional details are in the appendix.

\section{Limitations}
\label{sec:limitations}

\textbf{Quality.} Like many other image editing solutions, UniTune capabilities are tightly coupled to those of the underlying text-to-image model. Our choice in the experiments we've run was Imagen \cite{Imagen}, which is a robust high quality generator, especially for photo-realistic images. Nevertheless, our biggest losses are in the rare cases where Imagen faces difficulties. We are hopeful that as the base model improves so will UniTune here. Also, there are cases where subjects get mixed (e.g. the faces of two people are swapped) or cloned (e.g. the same face repeats more than once).
Showing the user several generations from different seeds usually mitigated all such issues, but there is more work for getting the system to output a single image that is always high quality.

\textbf{Balancing fidelity and expressiveness.} There are some instances where it was hard for us to find a good balance between fidelity and expressiveness, most notably when issuing small size edits, or when using edit masks (two cases that bias the model towards the base image). In those cases, the model transitions abruptly between copying the base image, and producing something too far from it. We believe improved sampling methods will help here, but leave exploration for future work. 

\textbf{Latency.} The first step of UniTune, fine tuning the base model, take around 3 minutes using TPUv4, and needs to be run once per input image. Then, generation with UniTune takes the same time as sampling from the original large model ($\sim$30 seconds). While workable, this is far from interactive. In addition, there is additional i/o cost (in memory or time) attributed to saving multiple copies of the fine-tuned model.

\section{Conclusion}
In this paper we presented UniTune, a simple and powerful approach for text-driven image editing. UniTune is unique in its abilities to make edits that require significant visual changes, as it can pick the correct visual and semantic details to preserve, only based on a textual description. This makes UniTune useful by casual users e.g. by speaking to a mobile device. We showed that fine-tuning a diffusion model on a single image is a promising method to bias its output distribution towards that image, and that, surprisingly, editing capabilities are preserved when using the right sampling methods.

Our work raises interesting questions more broadly, beyond image generation, on whether we could use similar techniques to imbue large models in other domains (e.g. text) with preferences by fine tuning on a single example.

Finally, there is a lot of room for further research, including how to better adjust the fidelity-expressiveness knobs, how to further increase the chances of a good single result, and how to improve generation speed. We believe that fine-tuned text-to-image diffusion models with the sampling mechanisms we propose are a good starting point for followup research.

\section*{Societal impact} UniTune, like other image generation models, has a great potential to complement and augment human creativity by creating new tools for professionals and empowering non-professionals with the ability to edit images more easily and in a more intuitive manner. However, we recognize that applications of this research may impact individuals and society in complex ways (see \cite{Imagen} for an overview). In particular, this method illustrates the ease with which such models can be used to alter sensitive characteristics such as skin color, age and gender. Although this has long been possible by means of image editing software, text-to-image models can make it easier.

Another cause of concern is reproducing unfair bias that may be found in the underlying model training data. This is also relevant for our underlying model, Imagen (see discussion in \cite{Imagen}). Moreover, these unfair biases may make the performance of the model vary across people of different groups. While we did not see this effect in our qualitative experiments, more research into bias evaluation methods, both for image editing and generation will help address this concern.

We encourage future research to help mitigate and measure the potential negative impact of generative models if misused, and believe thoughtful consideration and further research in all of these matters is necessary prior to determining how such technologies can be made broadly available.

\begin{acks}
We would like to express our gratitude to Mohammad Norouzi, William Chan, Yael Pritch, Daniel Cohen-Or, and Valerie Nygaard for their invaluable support throughout this project. We also thank numerous teams within Google Research, particularly the Imagen team, for  supporting this work. A special thank you goes out to our friends and family who generously provided photos for this paper. Human faces not belonging to friends and family were generated using StyleGAN \cite{StyleGAN}.
Furthermore, we extend our appreciation to Unsplash and the talented photographers Olia Gozha, Matthew Spiteri, Manja Vitolic, Luke Stackpoole, Andrea Lightfoot, and Justin Aikin for supplying all the other images used in this work as editing inputs (except those taken from other papers for comparison purposes, as noted in the relevant sections).
\end{acks}

\bibliographystyle{ACM-Reference-Format}
\bibliography{references}

\appendix

\section{Analysis of hyperparameters}
\Cref{fig:sdft_grid} offers a qualitative analysis of how the number of fine tuning iterations ("$FT$") and the initial sampling step ("$t_0$") affect UniTune output. The two parameters allow tuning between fidelity (faithfulness to the input photo) and expressiveness (faithfulness to the given edit prompt).  At every column expressiveness gets higher when going down the column, at a cost of a lower fidelity. As can be seen in the figure, fine-tuning on the input image (which corresponds to going right in the grid) allows us to reach a better mixture of fidelity and expressiveness.
To give a fairer chance for methods that mostly ignore the input image (e.g. high $t_0$ values), we use a very detailed edit prompt that describes the characters, clothing and poses. Otherwise, cells with no fine-tuning iterations and high $t_0$ values look very different from the original image.
Note that the figure can also serve as a qualitative comparison to SDEdit \cite{SDEdit}, as the first column ($FT=0$) is equivalent to not using any fine-tuning. The last row ($t_0=1$) is equivalent to not using any initialization and is therefore an ablation of that sampling technique.

\Cref{table:CatsParams} provides for all edit prompts and parameters used to generate the images in Fig. 9 in the main paper and can be used to see how these hyperparametrs are applied when generating images.

\begin{figure}[htbp]
\centering
\includegraphics[width=0.8\linewidth]{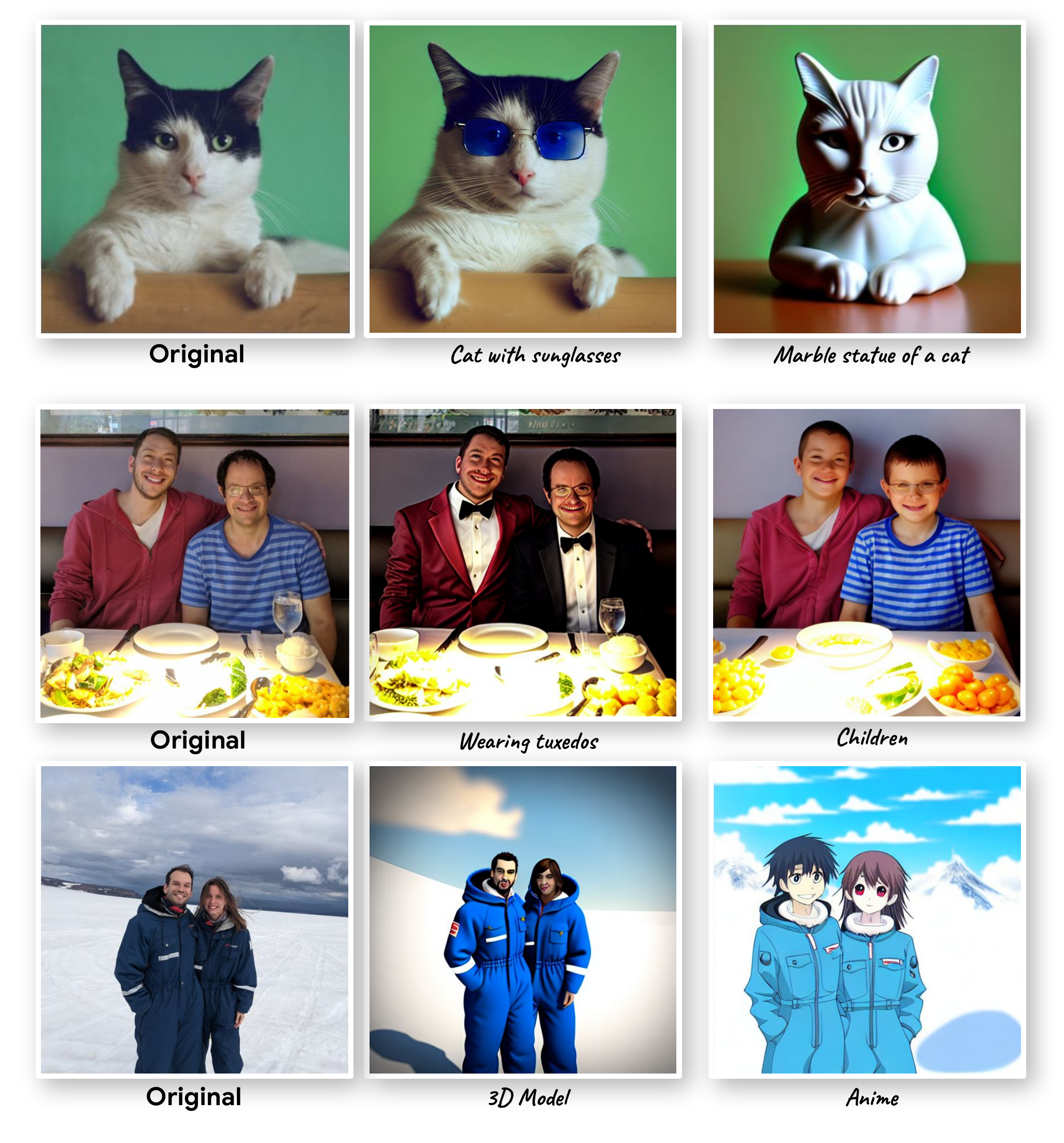}
\caption[test]{Selected UniTune samples with Stable Diffusion as the underlying model. Input source for row 1:  Unsplash.}
\label{fig:stable_diffusion}
\end{figure}

\begin{table*}[htbp]
\centering
\caption{Prompts and parameters used to generate the images in Fig. 9 in the main paper. All figures used a classifier free guidance weight of 32. Super-resolution model was not fine tuned for these images.}

\begin{tabular}{@{}lrr@{}}
\toprule
Edit Prompt                                       & Fine-Tuning Steps & Base Image Initialization \\ \midrule
Yellow flower                                     & 64                                    & 0.94                                      \\
A bee pollinating a flower                        & 64                                    & 0.90                                      \\
A ladybug on a flower                             & 64                                    & 0.88                                      \\
Black and white pencil sketch of a flower         & 32                                    & 1.00                                      \\
Elephants wearing top hats                        & 64                                    & 0.98                                      \\
Elephants walking in the water                    & 64                                    & 0.96                                      \\
Elephants walking in the thick jungle, many trees & 64                                    & 0.98                                      \\
Framed oil painting of elephants                  & 64                                    & 1.00                                      \\
A vampire cat wearing a coat                      & 64                                    & 0.94                                      \\
A cat peeking out of a basket                     & 64                                    & 0.98                                      \\
A cat wearing sunglasses                          & 64                                    & 0.90                                      \\
A brown cat with yellow eyes                      & 64                                    & 0.94                                      \\
A vampire cat wearing a coat, scary fangs         & 32                                    & 0.98                                      \\
Beautiful watercolor painting of a cat, on canvas & 32                                    & 1.00                                      \\
Birthday card, watercolor painting of a cat       & 32                                    & 1.00                                      \\
Marble statue of a cat                            & 64                                    & 0.90                                      \\ \bottomrule
\end{tabular}
\label{table:CatsParams}
\end{table*}

\begin{figure*}[htbp]
\centering
\includegraphics[width=0.85\linewidth]{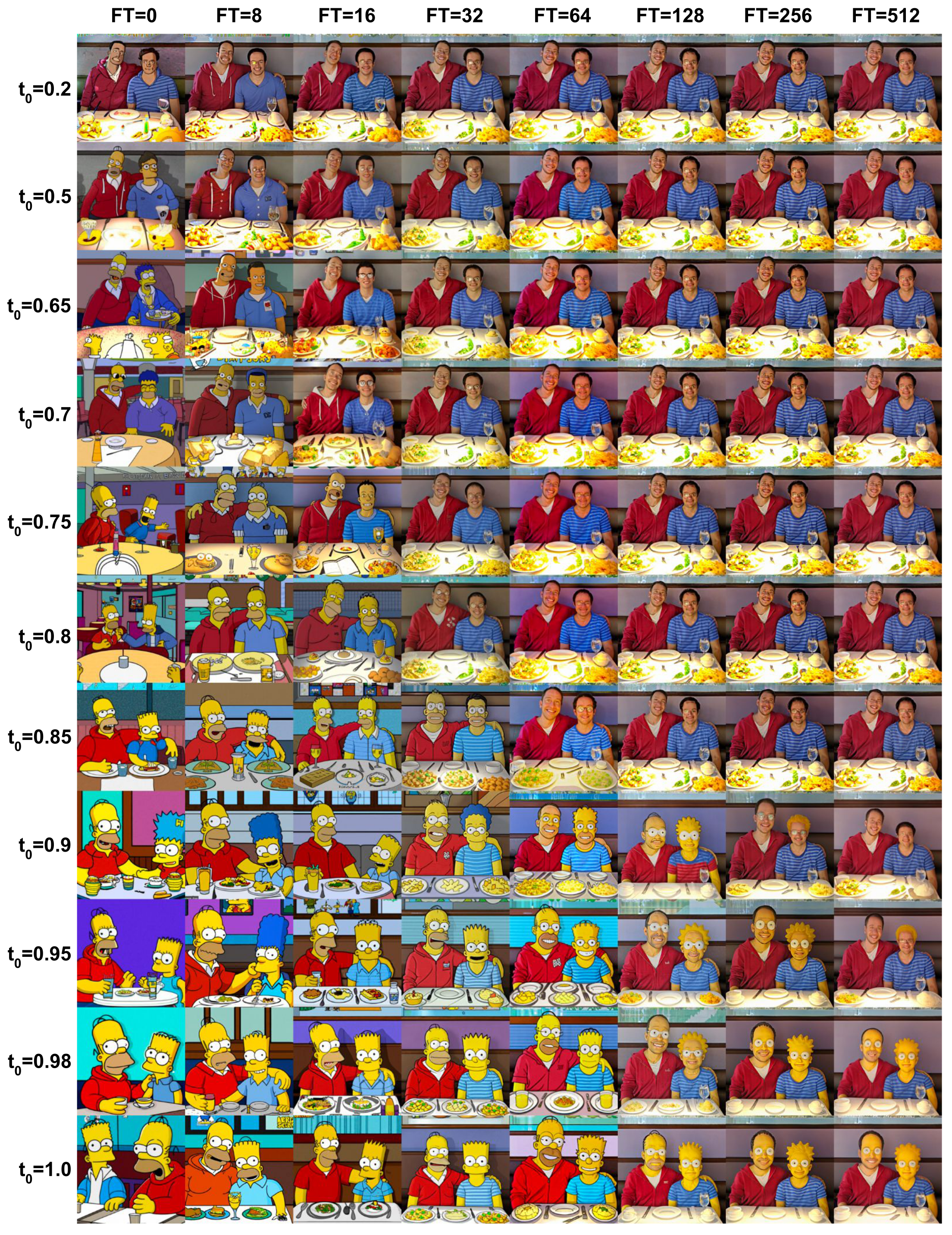}
\caption[test]{UniTune results for the same input image and edit prompt, with different combinations of fine-tuning iterations (x-axis) and initial sampling step (y-axis). The input image was the image on the left of Fig. 1 in the main paper. The edit prompt was "Homer and Bart Simpson sitting in a restaurant. Homer wears a red hoodie. Bart wears a blue shirt. Homer puts his hand on Bart's shoulder. drawn in the style of the Simpsons". We chose a very expressive query to maximize fidelity even for cases that ignore the input image (high $t_0$, no fine-tuning).}

\label{fig:sdft_grid}
\end{figure*}

\section{UniTune on Stable Diffusion}
UniTune's simplicity makes it easy to apply to different base architectures. While we haven't run rigorous testing, a simple port of our implementation of UniTune from Imagen to Stable Diffusion \citep{StableDiffusion} seems to generate similarly good results. Our Stable Diffusion implementation is almost identical to the implementation on top of Imagen, detailed in the main paper. Specifically, similarly to our Imagen implementation, our Stable Diffusion implementation uses Adafactor with a batch size of 4, emits weights at 16, 32, 64, 128 training steps, starts the sampling at step $0.8 \le t \le 0.98$ with appropriately noised image latents, and keeps Stable Diffusion's text encoder frozen.
The only modifications are using a learning rate of 2e-5 for fine tuning the model, and using a classifier free guidance weight of 7.5.
The weights of Stable Diffusion's VAE are kept frozen during fine tuning.
Selected samples from the Stable Diffusion implementation can be seen in
\cref{fig:stable_diffusion}.

\section{UniTune in-painting}
\label{sec:InPainting}
\begin{figure*}[htbp]
\centering
\includegraphics[width=0.8\linewidth]{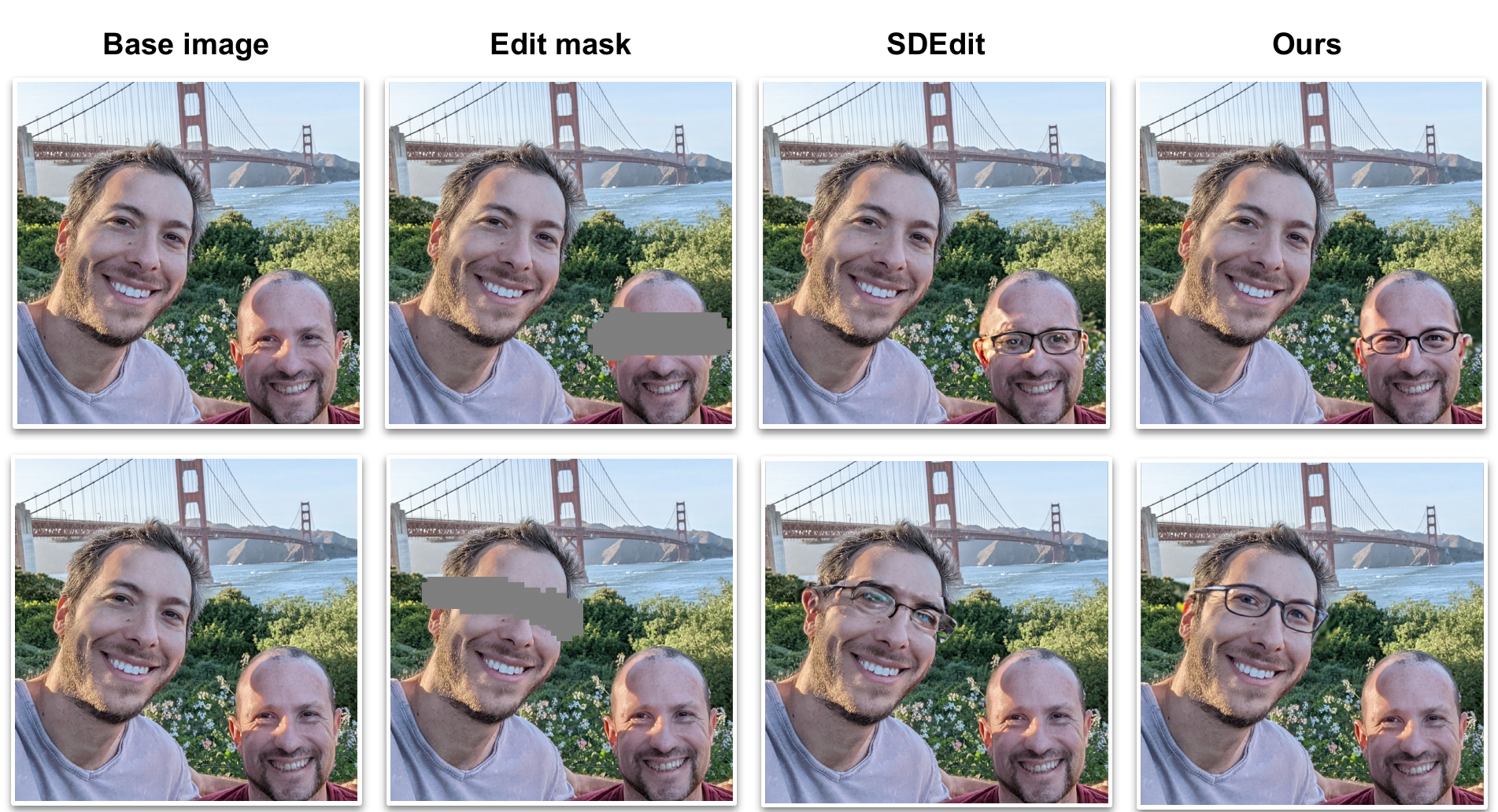}
\caption[test]{Adding glasses to faces with editing masks. UniTune is familiar with the details of the edited portion, and can use them when drawing the glasses to make the eyes more realistic and similar to the based image.}

\label{fig:sf_glasses}
\end{figure*}

\begin{figure*}[htbp]
\centering
\includegraphics[width=0.8\linewidth]{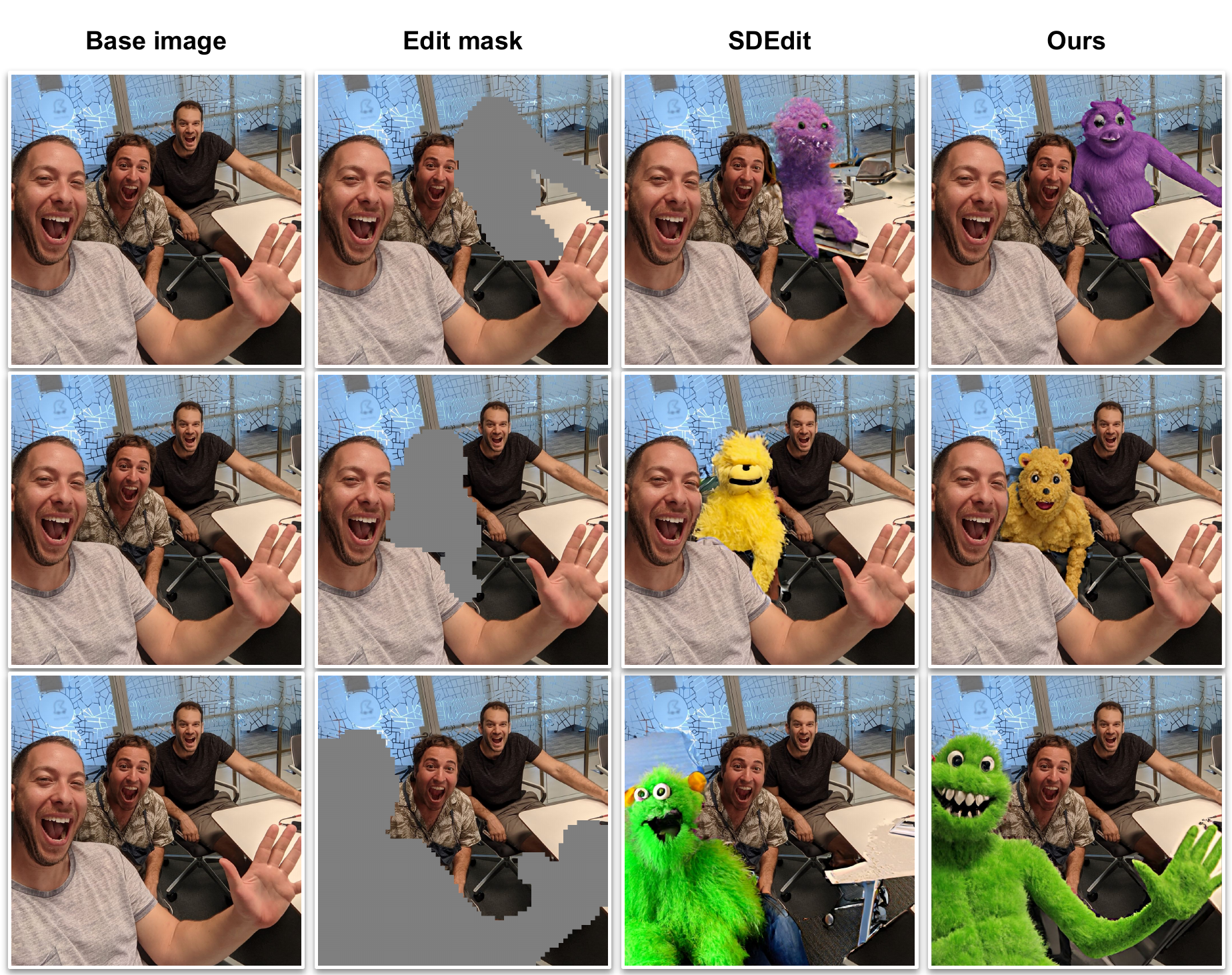}
\caption[test]{Turning people into fluffy monsters with editing masks. UniTune is able to maintain the posture and arm alignment even though it's hidden behind the edit mask.}
\label{fig:monsters}
\end{figure*}

Limiting edits to a predefined area ("edit mask") is a common technique to maintain fidelity to specific parts of the base image and to control edit sizes (at the price of expressiveness as it's often hard to predict the right proportions of the edited section). In diffusion models, this is often done by replacing a non-edited portion of the image with an appropriately noised version of the base image at every sampling step \cite{BlendedDiffusion}, or by overriding the model output at every step as if it predicted the original image outside the mask.

While UniTune works well without any masks, and is therefore usable by non-experts (e.g. on a mobile device by simply speaking commands), it is compatible with common masking techniques. UniTune is especially useful when a higher fidelity is needed within the masked area as it is familiar with the image details. This also allows imprecise large masks as UniTune can restore the unedited parts inside the mask correctly. \Cref{fig:sf_glasses} demonstrate how UniTune in-painting can be used when adding transparent elements like eyeglasses, and \cref{fig:monsters} shows how it can be used when editing existing characters or objects within a photo while maintaining semantic fidelity. Following \cite{BlendedDiffusion}, our implementation of UniTune in-painting replaces non-edited portions of the image with an appropriately noised version of the base image at every sampling step.

\begin{table*}
\centering
\caption{Categorization of the human evaluation set into 3 buckets, based on the initial CLIP score of the prompt vs. the base image. The UniTune preference column measures the ratio of ratings of (base image, edit prompt) pairs for which human raters preferred the best image generated by UniTune more than the best image generated by SDEdit. The last 3 columns measure the average scores of the winning images in each bucket. Norm-CLIP is the ratio between the CLIP score of the winning image (vs. the edit prompt) and the initial CLIP score of the base image (for the same prompt).}
\begin{tabular}{|c c c c c c c|} 
 \hline
 Bucket   & n &  Base image CLIP range    & UniTune preference    & Winner's MSE  & Winner's CLIP & Winner's Norm-CLIP \\
 \hline
1   & 31    & [0.081-0.167]   & 4.2   & 0.34   & 0.27  & 2.09 \\
2   & 32    & [0.168-0.241]   & 1.5   & 0.24   & 0.31  & 1.48 \\
3   & 31    & [0.243-0.327]   & 1.1   & 0.21   & 0.33  & 1.24 \\ 
 \hline
\end{tabular}
\label{table:RatingsByOrgClip}
\end{table*}

\section{Human evaluation details}

In this section we describe in detail the methodology for the quantitative comparison of UniTune and SDedit. 

\subsection {Dataset}

The evaluation was done on a dataset of 93 (base image, edit prompt) pairs with images of animals, people, objects, food and scenery. We designed the dataset such that half of the edits require significant pixel-level visual changes (like moving objects, zooming out or making significant style changes) and the other half only requires smaller local edits (like adding or replacing a small object, or changing styles but preserving overall coloration). 

For each (base image, edit prompt) we generate 13 SDEdit images and 13 UniTune images. For SDedit we generate images with different initial sampling steps $0.2\leq t_0\leq0.8$:
$\lbrace 0.20, 0.25, 0.30, ..., 0.80 \rbrace$. The $t_0$ range was selected to maximise the performance of the method. For UniTune, we generated images with the following combinations of fine-tuning iterations and initial sampling step:
\small
\linebreak
$(16, 1.0), (16, 0.85),(16,0.8), (32, 1.0), (32, 0.98), (32, 0.95), \linebreak 
(32, 0.9), (32,0.85), (32, 0.8), (64, 1.0), (64, 0.98), (64, 0.95), \linebreak
(128, 1.0)$. \normalsize For each of the 13 SDEdit and UniTune configurations we generated images from 8 random seeds using 256 DDPM steps, and picked the one with the lowest CLIP \cite{CLIP} distance to the edit prompt, resulting in a single image per configuration. We did not use super-resolution fine-tuning and interpolation.

\subsection {Evaluation process}
\begin{figure*}[p]
\centering
\includegraphics[width=1\linewidth]{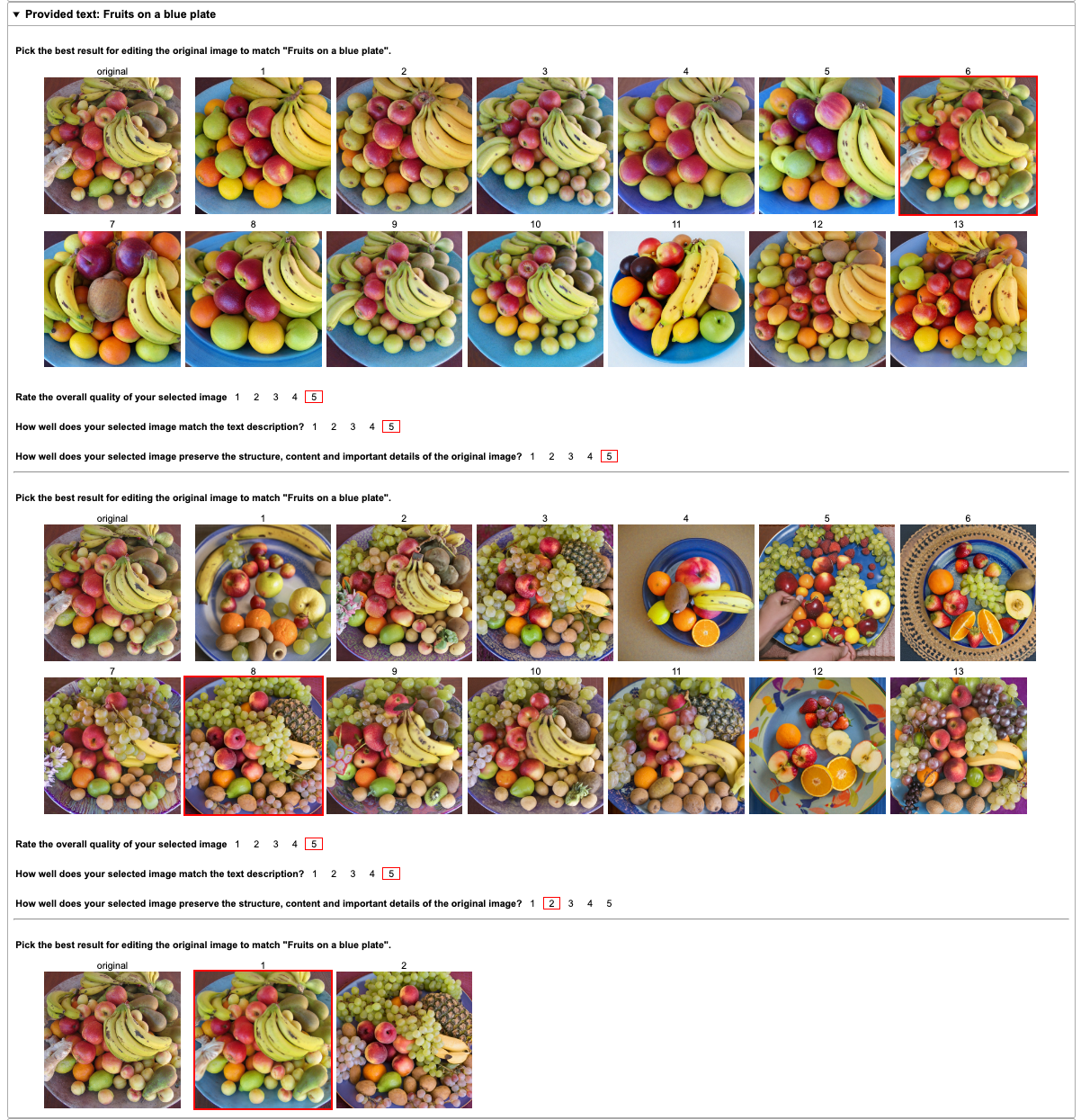}
\caption[test]{A screenshot of our evaluation app.}
\label{fig:eval-app-screenshot}
\end{figure*}

The evaluation was performed by a panel of 29 human raters. Raters were shown questions on pairs of (base image, edit prompt). For each pair and edit method (UniTune and SDEdit), the rater is shown a set of 13 edited images next to the base image, and has to pick the best one, balancing fidelity and adherence to the edit prompt. Next, the rater scores the chosen image on a scale of 1-5 on for quality, text alignment and fidelity (see \cref{fig:eval-app-screenshot} for the exact phrasing). Finally, the rater is shown the 2 chosen images next to the base image  and picks the best one. This determines the choice of the preferred method for the (base image, edit prompt) pair.

To ensure double blindness, we used random ordering of the (base image, edit prompt) pairs, the methods within each pair, and the 13 images belonging to a method. We collected 15-29 answers for each pair and weighted the responses appropriately.

\subsection {Additional analysis}
In Sec 5.1 in the main paper we showed that prompts that require significant changes vs. the base image benefit more from using UniTune vs. SDEdit by manually partitioning the evaluation dataset into the 'Aligned' and 'Unaligned' buckets. We see that UniTune's ratings are relatively consistent between the buckets, while SDEdit performs worse in the unaligned bucket with ratings going down by \texttildelow10\%. 

To estimate which edits require significant changes automatically, we can use CLIP \cite{CLIP} to measure the similarity of the edit prompt and the base image. Following the original CLIP training loss, we define CLIP score as $f(text) \cdot g(image)$ where $f$,$g$ are the CLIP text and image embeddings. Therefore, the lower the CLIP score of the edit prompt vs. the base image, the bigger the change that would be required. Note that this metric does not perfectly capture expected change in terms of MSE: changes that are large in terms of MSE may be small in terms of semantic distance (e.g. moving objects).

Based on the above, we partitioned the evaluation dataset into 3 buckets of equal size, and analyzed the human ratings for each of these buckets  (see \Cref{table:RatingsByOrgClip}). As expected, we observe that for (base image, edit prompt) pairs with the lowest CLIP scores, UniTune was preferred by the human raters 4.2:1 times, while in the second bucket it drops to 1.5:1, and for third bucket it reaches 1.1:1.

To further understand the differences between the buckets, we analyzed the winning images by each rater, for each bucket. We see that the winning image MSE is highly affected by the initial CLIP score bucket, suggesting that more pixels need to change in the base image to achieve good results. We also observe that the average CLIP scores for the winning images is lower for the buckets with lower initial CLIP scores, and we suspect the reason is that these edits are harder to achieve. We also computed a normalized CLIP score (Norm-CLIP), as the ratio between the CLIP score of the winning image vs. the initial CLIP score of the base image for the given edit prompt, to measure how big of a semantic change the model was required to achieve. As expected, we see that the norm-CLIP is significantly higher for the buckets with the lower CLIP scores. See all the details in \Cref{table:RatingsByOrgClip}.

\begin{figure*}[htbp]
\centering
\includegraphics[width=0.9\linewidth]{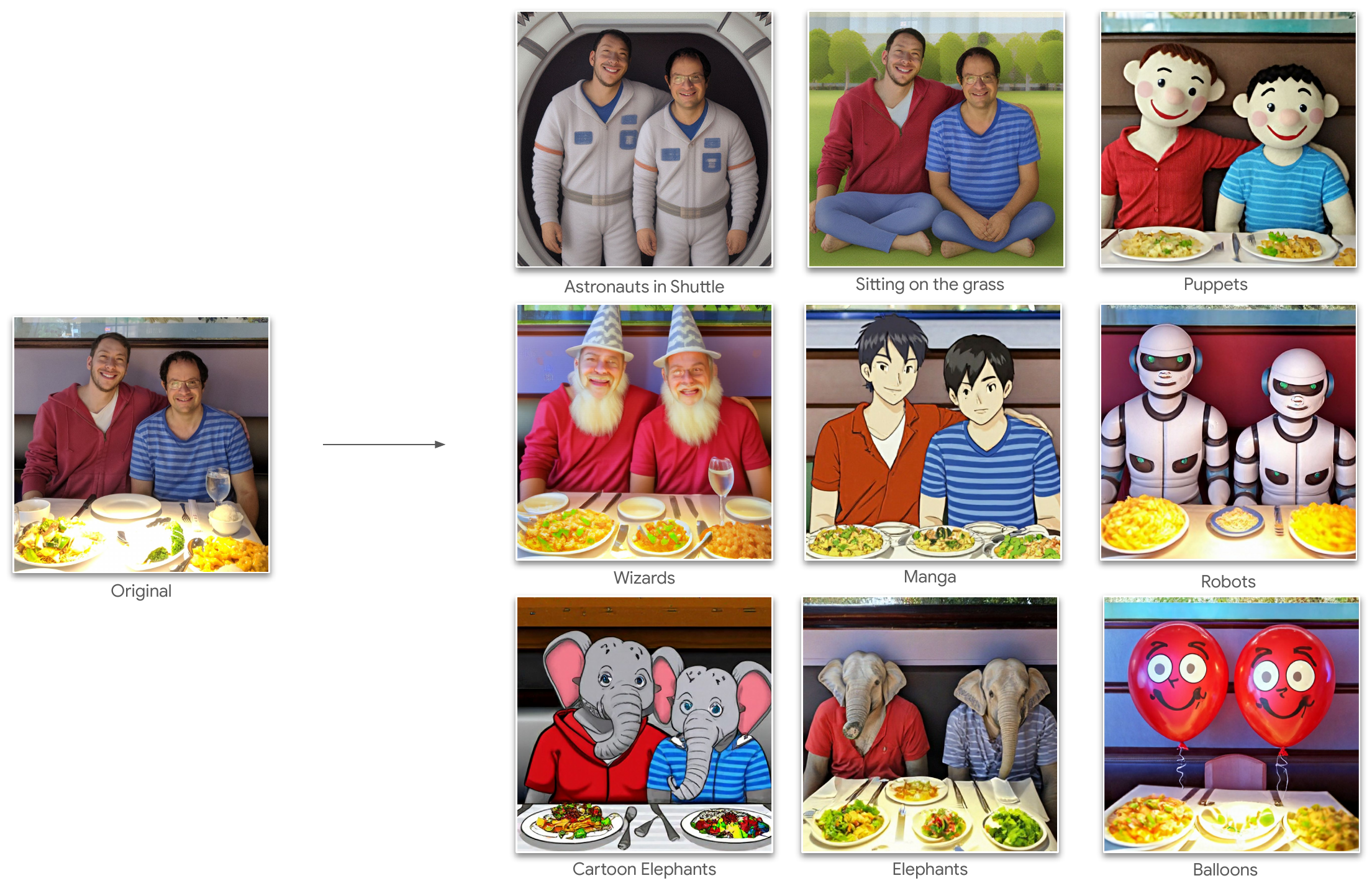}
\caption[test]{Additonal examples.}
\label{fig:edit_variety_more}
\end{figure*}

\begin{figure*}[htbp]
\centering
\includegraphics[width=0.75\linewidth]{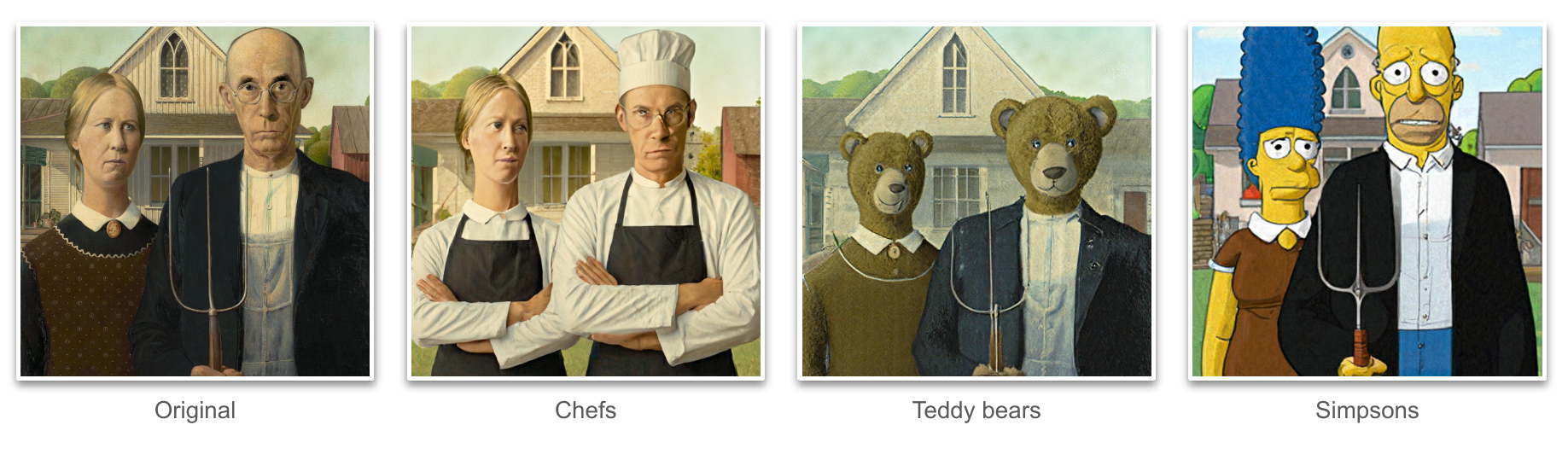}
\caption[test]{Transforming artworks using UniTune. Input source: Wood, Grant. American Gothic, 1930. © 2023 Figge Art Museum, successors to the Estate of Nan Wood Graham / Licensed by VAGA at Artists Rights Society (ARS), NY
}
\label{fig:american_gothic}
\end{figure*}

\end{document}